\documentclass[sigconf]{acmart}

\AtBeginDocument{%
  \providecommand\BibTeX{{%
    \normalfont B\kern-0.5em{\scshape i\kern-0.25em b}\kern-0.8em\TeX}}}

\copyrightyear{2022} 
\acmYear{2022} 
\setcopyright{acmlicensed}\acmConference[WWW '22 Companion]{Companion Proceedings of the Web Conference 2022}{April 25--29, 2022}{Virtual Event, Lyon, France}
\acmBooktitle{Companion Proceedings of the Web Conference 2022 (WWW '22 Companion), April 25--29, 2022, Virtual Event, Lyon, France}
\acmPrice{15.00}
\acmDOI{10.1145/3487553.3524729}
\acmISBN{978-1-4503-9130-6/22/04}

\usepackage{amsmath, multirow, subfigure, bm}

\begin{document}

\newtheorem{dfn}{Definition}

\title{Multi-Graph based Multi-Scenario Recommendation in Large-scale Online Video Services}


\author{Fan Zhang}
\orcid{0000-0001-5250-1323}
\affiliation{%
  \institution{OPPO Research Institute}
  \city{Shenzhen} 
  \country{China}
}
\email{zhangfan@oppo.com}

\author{Qiuying Peng}
\orcid{0000-0002-5373-0125}
\affiliation{%
  \institution{OPPO Research Institute}
  \city{Shenzhen} 
  \country{China}
}
\email{pengqiuying@oppo.com}
\authornote{Corresponding author}

\author{Yulin Wu}
\orcid{0000-0003-1077-9559}
\affiliation{%
  \institution{OPPO AI$\&$Data Engineering System}
  \city{Beijing} 
  \country{China}
}
\email{wuyulin@oppo.com}

\author{Zheng Pan}
\orcid{0000-0002-4487-3140}
\affiliation{%
  \institution{OPPO Research Institute}
  \city{Shenzhen}
  \country{China}
}
\email{panzheng@oppo.com}

\author{Rong Zeng}
\orcid{0000-0003-3741-0567}
\affiliation{%
 \institution{Bytedance Ltd.}
 \city{Shenzhen} 
 \country{China}
}
\email{zengrron@gmail.com}
\authornote{Work done when authors were interning at OPPO}

\author{Da Lin}
\orcid{0000-0003-1137-1803}
\affiliation{%
  \institution{Bytedance Ltd.}
  \city{Beijing} 
  \country{China}
}
\email{desmondl7@126.com}
\authornotemark[2]

\author{Yue Qi}
\orcid{0000-0003-1912-7418}
\affiliation{%
  \institution{OPPO Research Institute}
  \city{Shenzhen} 
  \country{China}
}
\email{qiyue@oppo.com}

\renewcommand{\shortauthors}{Zhang, et al.}

\begin{abstract}

  Recently, industrial recommendation services have been boosted by the continual upgrade of deep learning methods. However, they still face de-biasing challenges such as exposure bias and cold-start problem, where circulations of machine learning training on human interaction history leads algorithms to repeatedly suggest exposed items while ignoring less-active ones. Additional problems exist in multi-scenario platforms, e.g. appropriate data fusion from subsidiary scenarios, which we observe could be alleviated through graph structured data integration via message passing.

In this paper, we present a multi-graph structured multi-scenario recommendation solution, which encapsulates interaction data across scenarios with multi-graph and obtains representation via graph learning. Extensive offline and online experiments on real-world datasets are conducted where the proposed method demonstrates an increase of $0.63\%$ and $0.71\%$ in CTR and Video Views per capita on new users over deployed set of baselines and outperforms regular method in increasing the number of outer-scenario videos by $25\%$ and video watches by $116\%$, validating its superiority in activating cold videos and enriching target recommendation.

\end{abstract}

\ccsdesc[500]{Mathematics of computing~Graph algorithms}
\ccsdesc[500]{Information systems~Recommender systems}
\ccsdesc[500]{Computing methodologies~Learning latent representations}
\ccsdesc[500]{Computing methodologies~Neural networks}

\keywords{Graph Neural Networks, Multi-Scenario Recommendation, Recommender Systems}

\maketitle

\section{Introduction}

Online video recommendation services, as one specialization of recommender systems, have been playing an important role in internet entertainment. Deep learning models are widely deployed to engage users as much as possible, including collaborative filtering \citep{Cheng2016WideD, Covington2016DeepNN}, factorization machine \citep{Rendle2010FactorizationM, Ma2016FieldAwareFM, Guo2017DeepFMAF}, and also needs to be mentioned, graph embedding \citep{Wang2018BillionscaleCE} method.

\begin{figure}[htb]
  \centering
  \includegraphics[width=0.8\columnwidth]{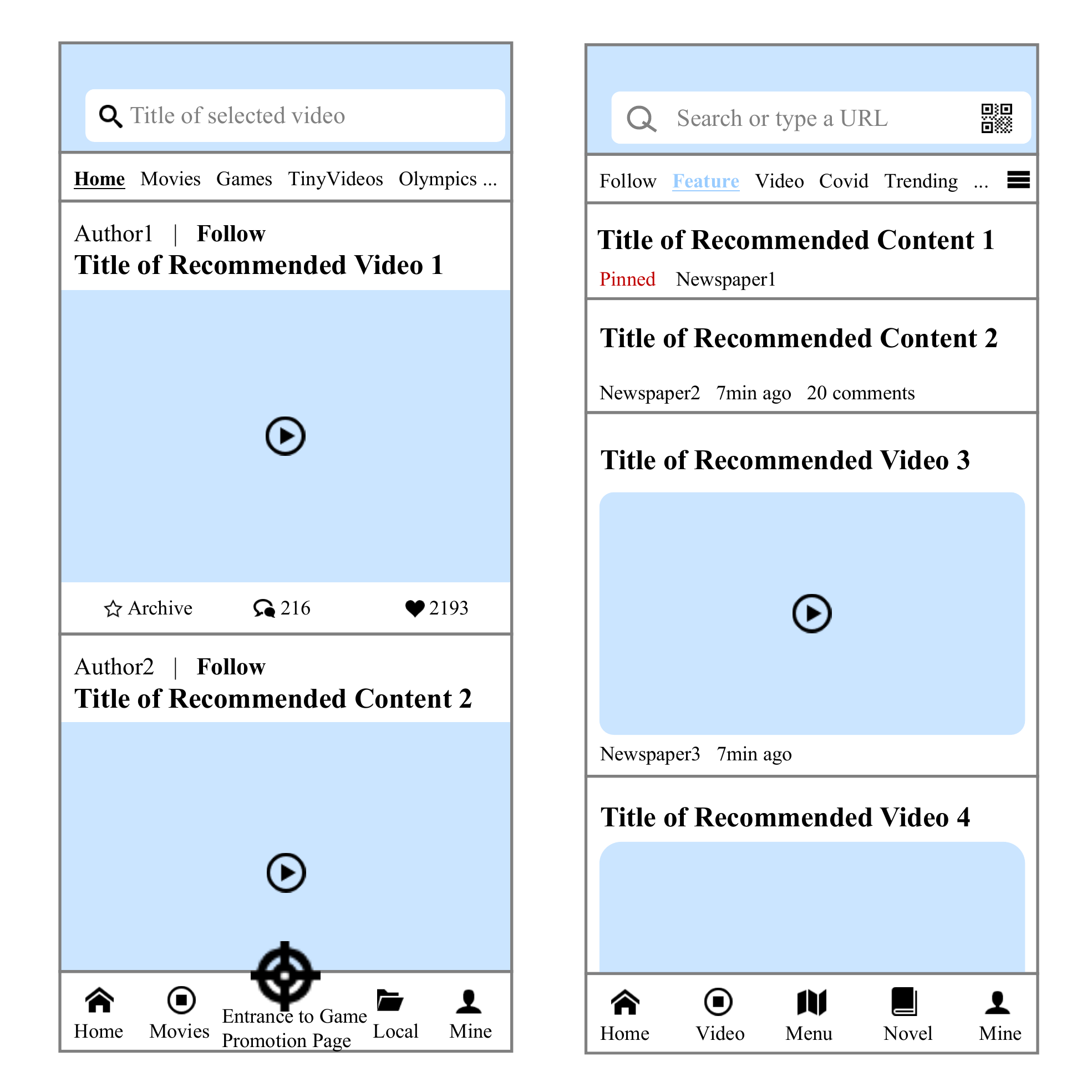}
  \caption{An example of online video recommendation scenarios: Homepage scenario of Shipin (left) and Feeds scenario of Browser (right). Shipin and Browser are two separate apps both featuring streaming services. Shipin only provides video watching services. Browser Feeds delivers both text- and video-format media services, where text- and video-format contents are suggested by separate retrieval algorithms. In addition, the video-format content from Shipin Homepage and Browser Feeds are supplied by a same third-party service.}
  \Description{Two images of user interface layout of Shipin Homepage scenario (left) and Browser Feeds scenario (right).}
  \label{TwoScenarioView}
\end{figure}

Vast majority of the industrial recommendation systems behind each \textit{scenario} are developed and maintained by individual groups of staff due to commercial concerns and dissimilar scenario characteristics, but there are some unnoticed issues underneath. Take the following two scenarios in Fig.~\ref{TwoScenarioView} as an example: 1) Shipin Homepage - the homepage recommendation scenario of Shipin, and 2) Browser Feeds - the personalized feeds recommendation service in Browser. Underlying issues of independent algorithm development include: 1) \textbf{Exposure bias}: due to the over abundant videos provided, users tend to select videos from exposed videos suggested by the system. At the end of the day, interaction feedback on exposed videos would again serve as positive records for next round of model training, thus resulting in a situation where positive records are always narrowed to the already exposed videos. Even with same video reservoir, after months of user feedback and algorithm selection, the recommended set of videos in two scenarios would both acceleratingly bias towards the user-selected videos in their own scenario, leading to a scenario-wise biased interaction pattern. 2) \textbf{Cold-start problem}: cold-start problem is a common challenge to all recommender systems that it's difficult to make recommendation on items with insufficient gathered information. But it could be the case when sometimes a less-watched video in one scenario is at the same time a trending video in another, which intrigues the idea of introducing videos across scenarios as a partial solution in multi-scenario setting.

Therefore, the multi-scenario recommendation problem comes to us: \textbf{Is there a method to incorporate interaction pattern from multiple scenarios with similar recommendation settings, so as to alleviate both exposure bias and cold-start problems in target scenario?} The method should preferably be done in the matching stage instead of ranking, such that more wide-ranging videos can be picked out during candidate retrieval. Item-to-item manner is preferred since it violates privacy policy to distribute user-sensitive data \textit{out of} individual apps (more discussions in \ref{privacyconcerns}). It is intuitive to associate with transfer learning, as proposed by Alibaba and JD \citep{Zhao2020CATNCR, He2020DADNNMC}, while we argue that major approaches of transfer learning are not applicable in our platform for the following reasons: 1) Sample selection prior to learning: importing cross-scenario data directly would be the easiest to implement, while practical experiments proved its limited effects, which we will discuss later. Feature alignment is infeasible here since videos share the same features across apps, and label alignment would rather result in loss of domain-variant information; 2) Knowledge transfer: domain adaptation method would fail due to invariance in interaction patterns resulted from exposure bias; 3) Multi-task learning: not applicable here since various scenarios are not trained simultaneously, and models are not shared. 

Therefore, we propose a novel graph-based i2i recommendation method to learn multi-scenario video representations with consideration to scenario-variant interaction pattern. More specifically, adjacent videos within a certain time period in user-video interaction history sequences usually indicate a continuous interest transition, therefore by integrating video-to-video transition pairs over all users' watch sequences, a global video interest transition graph can be constructed, whose nodes represent the set of videos and edges represent global interest connections between videos. Recent studies \citep{Ying2018GraphCN, Wang2018BillionscaleCE} have proposed various innovative graph learning methods based on this interest transition graph, and achieved significant effects in Pinterest and Alibaba. Considering the consistency in video features and a variety in transition patterns across scenarios, we propose \textbf{Cross-Scenario Multi-Graph (CSMG)} to depict the cross-scenario video transition graph. The advantages of this method are: 1) the structure of multi-graph is naturally able to formulate one type of videos with nodes for simplicity, and various types of transitions with edge attributes for distinction, and is easily extendable to more scenarios in the future; 2) the message passing mechanism allows for easy integration on scenario-wise interaction with graph learning modules. 

Based on CSMG, we further propose a novel model \textbf{Multi-Graph Fusion Networks (MGFN)}, whose graph convolution module captures scenario-wise pattern in each scenario-subgraph and fusion module stitches them together to obtain a global representation, along with a specific cross-scenario negative sampling strategy. In online experiments, MGFN shows an increase of $0.63\%$ and $0.71\%$ in CTR and Video Views per capita on new users over state-of-the-art baseline, and the amount of video watches imported from Browser to Shipin increases by $116\%$ when compared to regular data importation method, which validates MGFN's superiority in activating cold videos. The main contributions of our work are:
\begin{itemize}
\item{We formulate the essential multi-scenario recommendation problem for platforms serving multiple subsidiary scenarios.}
\item{We present a universal Cross-Scenario Multi-Graph to encapsulate multi-scenario interaction patterns, and provide detailed reasoning of the working mechanism behind the proposed multi-graph prototype.}
\item{We propose a novel Multi-Graph Fusion Networks to extract inner-scenario topological patterns and to assemble cross-scenario interaction information.}
\item{Extensive offline and online experiments are conducted on real-world datasets. Results validate the overall effectiveness of MGFN and its superiority in alleviating exposure bias and activating cold videos. To our best knowledge, it is the first graph-based cross-scenario recommendation solution which have been examined in commercial production setting.}
\end{itemize}

\section{Problem Formulation}

In this section, we first briefly introduce the backgrounds of Graph Neural Networks, then we elaborate on the construction of CSMG and how CSMG's composition contributes. 

\subsection{Intro to Graph Neural Networks}

The deep learning based Graph Neural Networks (GNN) is first introduced by GraphSAGE \citep{Hamilton2017InductiveRL} and its industrial practice PinSAGE \citep{Ying2018GraphCN}, which proposed the novel idea of learning nodes' local neighborhood aggregators as topological signal, thus flattening the message passing mechanism into stackable layers of learnable feature projectors and message aggregators. Given node $v$ with representation $h_v$, the $k$-th layer of a GNN is concluded exactly as in \citep{Xu2019HowPA}, 
\begin{eqnarray}
a_v^{(k)} &=& \mathrm{AGGREGATE}^{(k)} \left( \{ h_u^{(k-1)}: u \in \mathcal{N}(v) \} \right) \\
h_v^{(k)} &=& \mathrm{COMBINE}^{(k)} \left( h_v^{(k-1)}, a_v^{(k)} \right)
\end{eqnarray}

The inferred node embeddings could be applied in three types of downstream tasks: node classification, link prediction and graph-level classification. In most retrieval tasks which are shaped as link prediction problems, the node embeddings are used to predict probability of link existence and fed into pairwise loss for training.

\subsection{Construction of Cross-Scenario Multi-Graph}

The user records are collected real-time through server logs and event tracking. Due to occasional delays in data postback, duplicated records of user-video watches are removed. Records of users and videos having no registered data are filtered out. In addition, only records with a watch completion rate greater than $3\%$ are considered valid.

After obtaining user interaction history in a certain time window $[t_1, t_2]$, the item-item graph can be constructed as described in \citep{Wang2018BillionscaleCE}. For each user $u_i \in \mathcal{U}$ in scenario $s \in \mathcal{S}$, its watch history is sorted in descending order $\{v_1, v_2, \ldots, v_n \}$ where the latest watched video show up at the end of the sequence. Then we iterate over the sequence, and collect adjacent pairs of videos $(v_j, v_{j+1})$ whose time interval is shorter than $3600$ seconds. After processing all sequences in all scenarios, the stacked transition pairs of videos constructs a multi-scenario video-video transition multi-graph, whose nodes represent individual videos and edges $e_{ij}^s$ represent the total number of transitions from video $v_i$ to video $v_j$ in scenario $s$. In addition, the node features are preprocessed from selected static video characteristics: hashed keyword, hashed tag, hashed id, duration, degree. 

Hence, the definition of Cross-Scenario Multi-Graph is given as:

\begin{dfn}[Cross-Scenario Multi-Graph]
A directed multi-graph $\mathcal{G}(\mathcal{V}, \mathcal{E})$ with node set $\mathcal{V}$ of $N$ nodes, edge set $\mathcal{E}$ of $M$ edges, and node embedding $H \in \mathbb{R}^{N \times d}$. Edge attribute $e_{ij}^{s}$ is a scalar representing weight on the directed edge from $v_i$ to $v_j$ in scenario $s \in \mathcal{S}$. 
\end{dfn}

\subsection{Composition of Cross-Scenario Multi-Graph}

\begin{figure}[htb]
  \centering
  \includegraphics[width=\columnwidth]{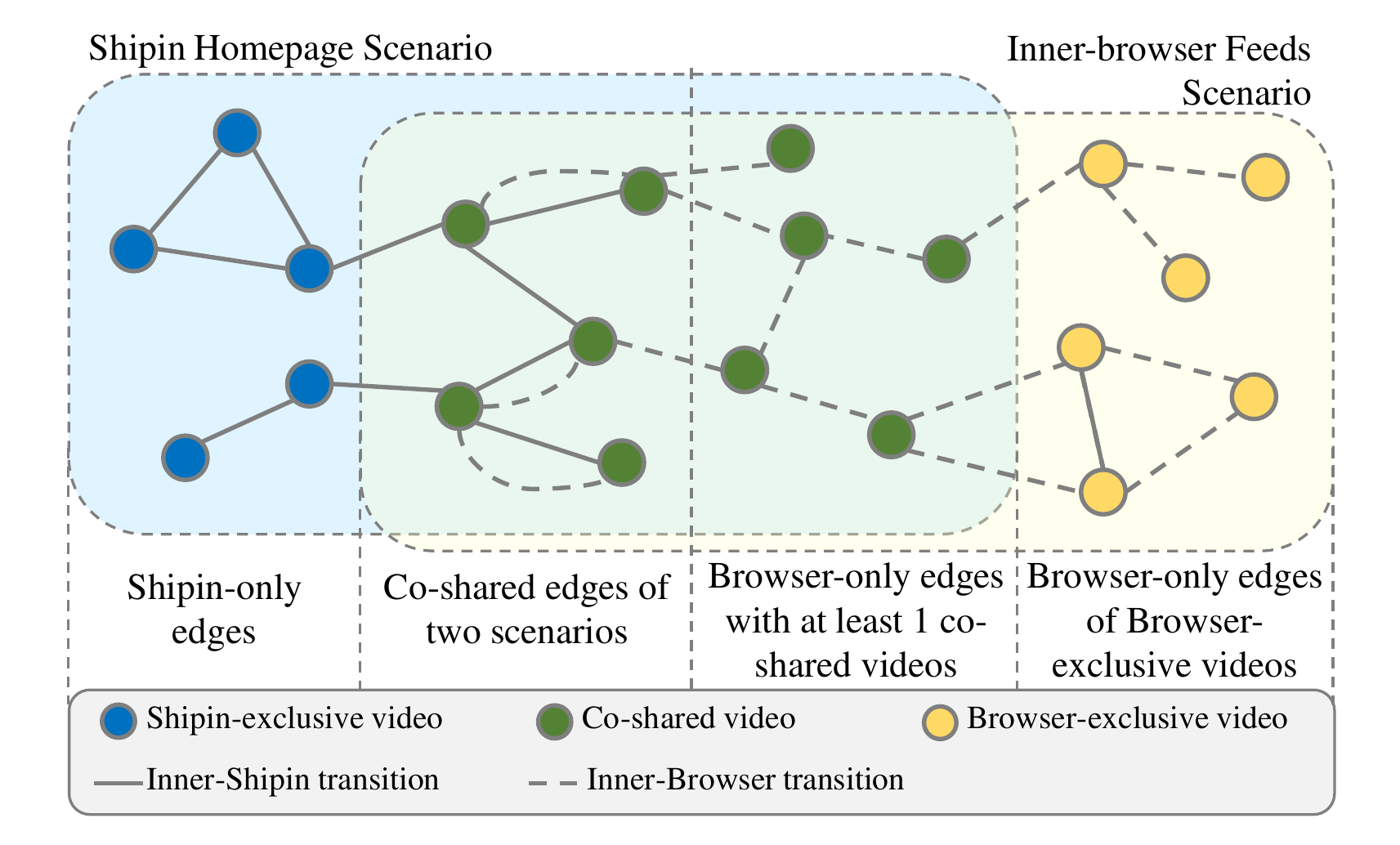}
  \caption{Edge composition of CSMG}
  \Description{There is a blue palette indicating nodes and edges in Shipin Homepage scenario, a yellow palette indicating those in Browser Feeds scenario, and the green-colored intersection of the two palettes indicates those overlapped nodes and edges. The green palette is aplit into two halves, where the left half includes co-shared edges of two scenarios and the right half includes Browser-only edges of at least 1 co-shared videos. The lines connecting nodes represent edges of cross-scenario multi-graph, where solid line represents inner-Shipin transitions and dashed line represents inner-Browser lines.}
  \label{edgetypeill}
\end{figure}

\begin{table}
  \caption{Edge composition of Cross-Scenario Multi-Graph. SE, SN denotes edge Existence or Non-existence in Shipin, and BE, BN denotes edge Existence or Non-existence in Browser.}
  \label{tab:edgecomptab}
  \begin{tabular}{p{0.3cm}|p{2.5cm}|p{4.5cm}}
     & SE & SN \\ \hline
    BE & Co-shared edges & Browser-only edges of Browser-exclusive videos + Browser-only edges with $\geq 1$ co-shared videos \\ \hline
    BN & Shipin-only edges & / \\
  \end{tabular}
\end{table}

Explorations on edge composition explains well about why CSMG is advantageous in multi-scenario recommendation. Specifically, we separate videos into three categories based on their appearances in the two scenarios (as shown in Fig.~\ref{edgetypeill}), where Shipin-exclusive videos represent videos which have only been watched in Shipin Homepage, Browser-exclusive videos represent those only been watched in Browser Feeds, and co-shared videos refer to those have been exposed and watched in both scenarios. Accordingly, we split the interest transition edges into $4$ groups based on their functionalities (Table~\ref{tab:edgecomptab}): 1) co-shared edges made of transition pairs which have appeared in both scenarios; 2) Shipin-only edges containing all transition pairs which have only appeared in Shipin; 3) Browser-only edges made of Browser-exclusive videos which have only appeared in Browser; 4) Browser-only edges with at least $1$ co-shared video on two ends, but have only appeared in Browser. 

While introducing videos from Browser (source) into Shipin scenario (target), the three types of edges from Browser could benefit the recommender in various aspects: 1) Co-shared edges: they usually compose of trending videos which have actively been watched in both apps, or are generated from strongly correlated videos. This type of edges is supposed to align node representation learning across scenarios, thus to guarantee a base performance for cross-scenario modeling. 2) Browser-only edges made of at least $1$ co-shared videos: the videos forming this type of edges have been exposed in Shipin scenario, while transitions between them haven't been captured in Shipin yet. However, since these transitions could appear in Browser, better capturing these transitions of video with multi-graph may supplement video global co-appearance and shorten their distance in the multi-scenario embedding space. This type of edges is the most important type of edges that we want our method to learn, and they may bring the most benefits. 3) Browser-only edges composed of Browser-exclusive videos: represent inner-Browser transition patterns. They exist in the multi-hop neighborhood away from Shipin-exclusive videos, and are connected to Shipin-exclusive videos through co-shared edges. The message passing mechanism of GNN allows for distant modeling for this type of edges, consequently expanding the receptive field of recommendation algorithms. 

\begin{table}
  \caption{Statistics of CSMG edge composition}
  \label{tab:edgecomp}
  \begin{tabular}{p{4.2cm}cc}
    \toprule
    Edge Type & Number & Percentage \\
    \midrule
    Co-shared edges of two scenarios & $1,512,167$ & $4.05\%$ \\
    Browser-only edges with $\geq 1$ co-shared videos & $26,890,282$ & $71.93\%$ \\
    Browser-only edges of Browser-exclusive videos & $8,980,060$ & $24.02\%$ \\
    \bottomrule
    $\#$ edges in Browser scenario & $37,382,509$ & $100\%$ \\
    ($\#$ edges in Shipin scenario) & ($47,408,383$) &  \\
    \bottomrule
  \end{tabular}
\end{table}

Take one specific time window as an example, the composition of mentioned edges from Browser scenario is listed in Table~\ref{tab:edgecomp}, indicating a great potential in cross-scenario modeling.

\section{Related Works}

In this section, we list several essential works which have inspired us in the area of graph neural networks, graph-based recommendation networks and multi-scenario recommendation.

\textbf{Graph Neural Networks} Since the groundbreaking proposal of graph embedding networks Deepwalk \citep{Perozzi2014DeepWalkOL} and LINE \citep{Tang2015LINELI}, much attention have been brought to graph-structured pattern recognition. Afterwards, GCN \citep{Kipf2017SemiSupervisedCW}, GraphSAGE \citep{Hamilton2017InductiveRL}, GAT \citep{Velickovic2018GraphAN} have been proposed, establishing the foundation for graph learning. Besides, SIGN \citep{Rossi2020SIGNSI} and HAN \citep{Wang2019HeterogeneousGA} extended graph types from homogeneous to heterogeneous, STGCN \citep{Yu2018SpatioTemporalGC} and EvolveGCN \citep{Pareja2020EvolveGCNEG} attempted to merge time series with dynamic learning of emerging graphs.

\textbf{Graph Based Recommendation} Various popular recommendation algorithms are built upon user-item interaction data to encode user/item embeddings, so as to predict user preferences from embeddings \citep{Cheng2016WideD, Guo2017DeepFMAF, Covington2016DeepNN}. The intrinsic similarity between bipartite-structured graphs and user-item interaction data format motivates the intuitive idea of GNN recommenders. Influential examples include PinSAGE \citep{Ying2018GraphCN} which mines pin-board interaction, EGES \citep{Wang2018BillionscaleCE} who brings side info to item graphs. Later models made progressive improvements by finer modeling \citep{Wu2019SessionbasedRW, Berg2017GraphCM, Liu2020GraphNN} and introduction of Knowledge Graphs \citep{Wang2019KnowledgeGC, Wang2019KGATKG}, and expanded downstream applications to larger scope, e.g. friend suggestion \citep{Fan2019GraphNN, grafrank}, spam review detection \citep{Li2019SpamRD} and traffic forecasting \citep{Yu2018SpatioTemporalGC}.

\textbf{Multi-Scenario Learning} Several recent studies brought new insight into cross-scenario recommendation. MA-RDPG \citep{Feng2018LearningTC} proposed a multi-agent reinforcement learning framework, with an emphasis on cooperating scenarios with competitive relationship. While CATN \citep{Zhao2020CATNCR}, DADNN \citep{He2020DADNNMC}, NATR \citep{Gao2019CrossdomainRW} relied on transfer learning to conduct knowledge transfer while maintaining scenario-aware variances. SAML \citep{Chen2020ScenarioawareAM} adopted a mutual learning method for simultaneous training of multiple scenarios. DMGE \citep{Ouyang2019LearningCR} proposed a multi-graph to enclose user behaviors from multiple domains, and optimized its GNN through multi-objective gradient descent. 

\textbf{Comparison to DMGE \citep{Ouyang2019LearningCR}} During paper writing, we noticed the same idea of cross-scenario multi-graph already proposed in \citep{Ouyang2019LearningCR}. However after careful examination, we found a major difference in task objective, model training and experiments. \citep{Ouyang2019LearningCR} shaped its problem as multi-task learning, optimized its GNN with multi-objective gradient descent and verified offline performance in yielding mutual metrics gain. While our work focuses on enriching item variety in target scenario with explainable mechanism, trained with cross-scenario negative sampling and verified efficacy through online experiments. Mutual learning is infeasible in our online serving systems since online systems of two scenarios are isolated. Therefore, the novelty and usefulness of our work still provide fruitful insights to practitioners in the field.

\section{Model}

\begin{figure}[htb]
  \centering
  \includegraphics[width=0.8\columnwidth]{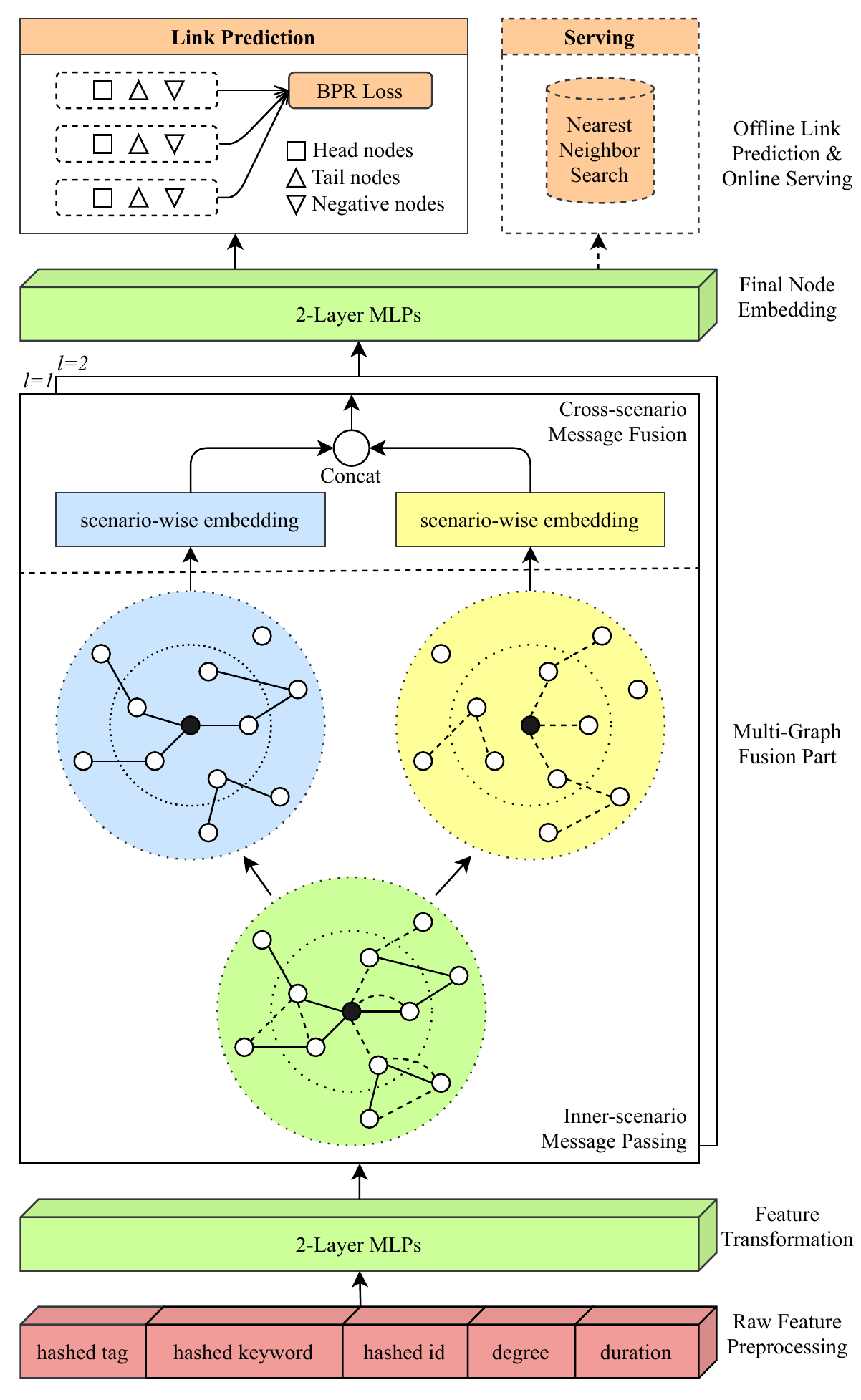}
  \caption{Overall architecture of MGFN.}
  \Description{The overall architecture, from bottom to top, is composed of raw feature processing module, feature transformation module, multi-graph fusion module, final node embedding, and offline link prediction and online serving module. The raw feature transformation is made of five blocks, from left to right, hashed tag, hashed keyword, hashed id, degree, duration. In the multi-graph fusion module, one green multi-graph circle points to one blue-colored solid-line circle and one yellow-colored dashed-line circle, and the two circles points to scenario-wise embedding separately and further merged towards the final node embedding module. The link prediction module is summed via BPR loss, and online serving module is supported by nearest neighbor search.}
  \label{overallarchitecture}
\end{figure}

The overall architecture of Multi-Graph Fusion Networks is illustrated in Fig.~\ref{overallarchitecture}. MGFN consists of three parts: the feature transformation part transforms video feature $H$ to expand the embedding space; the multi-graph fusion part extracts topological patterns; the link prediction part measures pairwise loss of triplets.

\subsection{Feature Transformation}

In order to convert the initial node embedding $H \in \mathbb{R}^{N \times d}$ to certain length $d_{0}$ for easier calculation, a Multi-Layer Perceptions (MLPs) is adopted. For node $v_i \in \mathcal{V}$, its embedding $h_i^{(0)}$ is initialized as:

\begin{equation}
h_i^{(0)} = W_{m_2}^T \cdot \sigma (W_{m_1}^T \cdot h_i + b_{m_1}) + b_{m_2}
\end{equation}

where $W_{m_1}, W_{m_2}, b_{m_1}, b_{m_2}$ denotes weight and bias terms of the $2$-layer MLPs, and $\sigma$ denotes non-linear activation unit ReLU.

\subsection{Multi-graph Fusion}

The multi-graph fusion part is designed to perform inner-scenario message passing via graph convolution modules to aggregate inner-scenario neighborhood message, and to conduct cross-scenario message fusion via various types of aggregation function so as to merge learnt video embeddings from scenarios.

\textbf{Inner-Scenario Message Passing} For each scenario $s \in \mathcal{S}$, the scenario-wise subgraph $G_s$ of nodes and edges is extracted from the original multi-graph $G$. Since an edge $e_{ji}^s$ represent an interest flow from $v_j$ to $v_i$ in scenario $s$, then for each node $v_i$ and its neighbor set $\mathcal{V}_s(i)$ in scenario $s$, the in-flowing neighbor message is considered valuable and similar to node embedding $h_i$. Here, graph convolution modules (e.g. SAGE, GAT) are strong candidates for propagating neighbor message to node $v_i$.

The message passed along edge $e_{ji}^s$ from neighboring node $v_j \in \mathcal{V}_s(i)$ to node $v_i$ in the $l$-th graph convolution layer is 
\begin{equation}   
m_{v_j \rightarrow v_i}^{s,l} = W_{\mathrm{nb}, l}^s \cdot e_{ji}^s \cdot  h_j^{(l-1)}
\end{equation}
which is scaled from neighbor $v_j$'s embedding $h_j^{(l-1)}$ with conversion matrix $W_{\mathrm{nb},l}^s$. Under the message passing mechanism, the message passed from a set of sampled neighbors is usually aggregated through an aggregator $\mathrm{AGG}$, and further combined with $W_{\mathrm{self}, l}^s$-scaled self embedding $h_i^{(l-1)}$ as 
\begin{equation}
h_{i, s}^{(l)} = W_{\mathrm{self}, l}^s \cdot h_i^{(l-1)} + \mathrm{AGG} \left(\{ m_{v_j \rightarrow v_i}^{s,l} \vert j \in \mathcal{V}_s(i) \} \right)
\end{equation}
The aggregator AGG varies with graph convolution modules (e.g. SAGE, GAT, GIN). Here we adopt SAGE-mean aggregator and GAT based on practical experiences. In a GraphSAGE prototype ("SAGE"), the inductive graph neural networks is stacked from learnable layers of graph convolution modules, which are made of an element-wise mean-pooling aggregator $\mathrm{AGG}$ of neighbor message and a learnable matrix $W_\mathrm{self}$ of node's self message. The element-wise mean pooling aggregator is defined as  
\begin{equation}
\mathrm{AGG}_{\mathrm{SAGE-mean}} = \frac{1}{\vert \mathcal{V}_s(i) \vert} \sum_{j \in \mathcal{V}_s(i)} m_{v_j \rightarrow v_i}^s
\end{equation}
Hence the $l$-th layer node embedding $h_i^{(l)}$ is obtained as
\begin{equation}
\mathrm{SAGE}: h_{i, s}^{(l)} = W_{\mathrm{self}, l}^s \cdot h_i^{(l-1)} + \frac{1}{\vert \mathcal{V}_s(i) \vert} \sum_{j \in \mathcal{V}_s(i)} W_{\mathrm{nb}, l}^s \cdot e_{ji}^s \cdot  h_j^{(l-1)}
\end{equation}
In a GAT prototype ("GAT"), a self-learnable attention weight $\alpha_{ji}^s$ is applied in aggregating neighbor message and is followed by $\sigma_{\mathrm{GAT}}$, a GAT-specific activation unit LeakyReLU, as given by 
\begin{equation}
\mathrm{AGG}_{\mathrm{GAT}} = \sigma_{\mathrm{GAT}} \left( \sum_{j \in \mathcal{V}_s(i)} \alpha_{ji}^s \cdot W_{\mathrm{nb}, l} \cdot e_{ji}^s \cdot h_j^{(l-1)} \right)
\end{equation}
where attention mechanism is performed with a shared attention $a : \mathbb{R}^d \times \mathbb{R}^d \rightarrow \mathbb{R}$ and a shared weight matrix $W_\mathrm{nb, l} \in \mathbb{R}^d$ as
\begin{eqnarray}
a_{ji}^{l,s} &=& a(W_\mathrm{nb, l} \cdot e_{ji}^s \cdot h_j^{(l-1)} , W_\mathrm{nb, l} \cdot e_{ji}^s \cdot h_i^{(l-1)}) \\
\alpha_{ji}^{l,s} &=& \mathrm{softmax}_j (a_{ji}^{l,s})
\end{eqnarray}
Here, the GAT module is followed by a skip-connection \citep{Xu2019HowPA} to boost its performance otherwise it faces severe underfitting in practice. We suspect that the intrinsic signal in node features is already strong enough and indisposable, that it'd be better to use GAT aggregation of neighbor message as a \textit{residual block} \citep{He2016DeepRL}. Therefore, the GAT module is given by:
\begin{equation}
\mathrm{GAT}: h_{i, s}^{(l)} = W_{\mathrm{self}, l}^s \cdot h_i^{(l-1)} + \sigma_{\mathrm{GAT}} \left( \sum_{j \in \mathcal{V}_s(i)} \alpha_{ji}^s \cdot W_{\mathrm{nb}, l} \cdot e_{ji}^s \cdot h_j^{(l-1)} \right)
\end{equation}
Note that in the case when node $v_i$ has no in-flow neighbor, its latent feature $h_{i,s}^{(l)}$ is learnt directly from its own embedding $h_i^{(l-1)}$. 

\textbf{Cross-Scenario Message Fusion} After obtaining scenario-wise latent vectors $h_{i,s}^{(l)}$ for node $v_i$, the cross-scenario combined latent feature $h_{i}^{(l)}$ at $l$-th layer can be leant via several types of fusion functions. Valid options include mean-pooling ("mean"), learnable weighted-average ("weighted") and concatenation ("concat"):
\begin{eqnarray}
\mathrm{mean}&:& h_i^{(l)} = \frac{1}{\vert \mathcal{S} \vert} \sum_{s \in \mathcal{S}} h_{i, s}^{(l)} \\
\mathrm{weighted}&:& h_i^{(l)} = \frac{1}{\vert \mathcal{S} \vert} \sum_{s \in \mathcal{S}} w_{s}^{(l)} \cdot h_{i, s}^{(l)} \\
\mathrm{concat}&:& h_i^{(l)} = h_{i, s=1}^{(l)} \mathbin\Vert h_{i, s=2}^{(l)} \mathbin\Vert \cdots \mathbin\Vert h_{i, s=S}^{(l)}
\end{eqnarray}

Finally, an extra $2$-layer MLPs is applied
\begin{equation}
h_i = W_{f_2}^T \cdot \sigma (W_{f_1}^T \cdot h_i^{(L)} + b_{f_1} ) + b_{f_2}
\end{equation}
to generate the final node embedding $h_i$ with learn-able weight and bias terms $W_{f_2}, W_{f_1}, b_{f_1}, b_{f_2}$. It eventually generalizes topologically-aggregated node embeddings into global video representations.

\subsection{Link Prediction and Training Process}

In retrieval tasks, pairwise loss (e.g. BPR loss, Hinge loss) is a popular method to evaluate model preference based on the assumption that users favor positive-labeled item pairs over negative-labeled item pairs. Here, we use BPR loss \citep{Rendle2009BPRBP} to evaluate the loss. For a given triplet $(v_h, v_t, v_n)$ where $(v_h, v_t)$ represents a positive transition pair and $(v_h, v_n)$ represents a non-existent negative transition pair selected with negative sampling strategy, the loss of a minibatch of $b_s$ triplets is calculated as follows,

\begin{equation}
\mathrm{Loss} = \frac{1}{b_s} \sum_{n=1}^{b_s} \log(1 + \exp(h_{\mathrm{v_h}} \cdot h_{\mathrm{v_n}} - h_{\mathrm{v_h}} \cdot h_{\mathrm{v_t}}))
\end{equation}
where, $h_{\mathrm{v_h}}$, $h_{\mathrm{v_t}}$, $h_{\mathrm{v_n}}$ represent the corresponding node embeddings, and $b_s$ is the number of $v_h$ nodes in each sampled minibatch.

The loss is minimized by an Adam optimizer with gradient descent until a certain stop step. The dataset is trained in an unsupervised manner with no cross validation, so overfitting is prevented via Dropout though offline experiments indicate optimal model performance with Dropout set to $0.0$.

\textbf{Cross-Scenario Negative Sampling Strategy} The positive item pairs are sampled from existed video transition pairs. Each positive item pair is matched with $5$ negative sampled item pairs. In addition, the negative items $v_n$ from negative item pairs $(v_h, v_n)$ are sampled according to the \textit{cross-scenario negative sampling strategy}, where the probability of a node $v_n$ being sampled $P(v_n)$ is proportional to the summation of its degrees in all scenarios $\sum_{s \in \mathcal{S}} \mathrm{deg}_s(v_n)$ raised to the $3/4$rd power \citep{Mikolov2013EfficientEO, Mikolov2013DistributedRO, Wang2018BillionscaleCE},
\begin{equation}
P(v_n) \propto \left( \sum_{s \in \mathcal{S}} \mathrm{deg}_s(v_n) \right)^{\frac{3}{4}}
\end{equation}
This negative sampling strategy is adopted to prevent over-smoothing on hot videos and to encourage video similarity learning closer to less-watched videos in all scenarios. In addition, random negative sampling is not adopted for the same reason to avoid leaning towards hot videos, and hard negatives (e.g. sampling videos inversely proportional to their average watch time) have been practised while online experiments show its limited effects. Besides, degree negative sampling in one single scenario would reduce model performance since it acts as random negative sampling in all other scenarios.

\section{Experiments}

\subsection{Experiment Setup}

\begin{table}
  \caption{Dataset statistics}
  \label{tab:datasetstats}
  \begin{tabular}{l|cc}
    \toprule
    Dataset & Commercial $1$ & Commercial $2$ \\
    \midrule
    \# Nodes & $818,551$ & $796,985$ \\
    \# Edges & $94,714,404$ & $73,210,932$ \\
    \# Edges (Shipin Homepage) & $47,408,383$ & $47,306,021$ \\
    \# Edges (Browser Feeds) & $37,382,509$ & $35,828,423$ \\
    \bottomrule
  \end{tabular}
\end{table}

\textbf{Datasets} For offline evaluation, we construct two private datasets from user interaction data collected in two real-world commercial apps, and use their next-day watch record in Shipin Homepage as corresponding offline validation dataset. Two datasets are selected with different contiguous $7$-day time window in June and July $2021$. Statistics of the two datasets is listed in Table~\ref{tab:datasetstats}.

\textbf{Offline Evaluation Metrics} For each user $u \in \mathcal{U}$ who have watched at least one video in both training and validation dataset, we compute the hits rate of the top-$100$ retrieved candidates (retrieved via same \textit{online retrieval strategy}) with its true watch history. Averaged over all users, the offline evaluation metrics $\mathrm{precision}@100 = \frac{1}{\vert \mathcal{U} \vert} \sum_{u \in \mathcal{U}} \frac{\mathrm{\# \, hits} (u)}{100}$ and $\mathrm{recall}@100 = \frac{1}{\vert \mathcal{U} \vert} \sum_{u \in \mathcal{U}} \frac{\mathrm{\# \, hits} (u)}{\mathrm{\# \, true \, watches}(u)}$ are  reported. Practical experiences indicate a positive relation between offline metrics and online performance.

\textbf{Online Retrieval Strategy} Video embeddings are computed offline daily based on the previous $7$-day window of user-video interaction data, and then transmitted to online serving system where all video embeddings are stored with Faiss \cite{JDH17} for fast similarity lookup. For each specific user, a queue of $20$ most recently viewed videos is updated real-time. Once a request is sent to server, each most recently viewed video in the queue will be matched to $15$ most similar videos with least inner product similarity. Among these $300$ matched videos, videos in collision with user watch history are removed, and the rest top-$100$ videos with highest ranking scores are passed forward to the ranking model. Same retrieval strategy is adopted offline in generating top-$k$ candidates. The online MGFN model is trained on a daily basis, which makes it less effective than online learning models in capturing real-time user interests, yet online learning of graph neural networks is another topic that we will not discuss in this paper. 

\textbf{Online metrics} The online metrics is set in accordance with the business goals of Shipin Homepage, here we use Click-Through-Rate (CTR), Video Views per capita (VV) and Video Watch Time per capita (VWT). CTR reflects a global user preference over exposed videos. VV and VWT are calculated as the average number of clicks users make and the average watch time users spend in the app, they can help quantize user stickiness to the app. A higher metrics value indicates a better real performance, and an increase rate of $0.5\%$ over "base" indicates a valid improvement ("base" refers to the already deployed set of retrieval models).

\subsection{Baselines}

We compare MGFN to: 1) model trained on Browser data for a validity test on exposure bias, 2) models trained on Shipin data as single-scenario graph recommenders, 3) model trained with naive multi-scenario approach, 4) variants of MGFN for ablation. The approaches for comparison are described as follows:
\begin{itemize}
\item \textbf{SAGE-Browser \citep{Hamilton2017InductiveRL}} is a GraphSAGE prototyped GNN built on video transition graph constructed from Browser Feeds data. Due to exposure bias, the model performs poorly in predicting a user's next-watch in Shipin Homepage.
\item \textbf{SAGE-Shipin \citep{Hamilton2017InductiveRL}} is a SAGE prototype built on Shipin Homepage data, and is conducting A/B testing at the same time.
\item \textbf{GAT-Shipin \citep{Velickovic2018GraphAN}} utilizes GAT instead of SAGE module in modeling Shipin Homepage data.
\item \textbf{DataConcat (Data Concatenation)} is a multi-scenario data importation method where user-video interaction data from multiple scenarios are simply concatenated into one. In this case, multiple scenarios are considered as one, and GraphSAGE is further applied to generate all video embeddings.
\item \textbf{MGFN-mean} is a variant of MGFN which uses mean-pooling instead of concatenation at cross-scenario fusion step.
\item \textbf{MGFN-weighted} is a variant of MGFN which uses weighted-average operation at cross-scenario fusion step.
\item \textbf{MGFN-GAT} is a variant of MGFN which uses GAT module (with skip connection) instead of SAGE module. 
\end{itemize}

Note that MGFN is not compared to SAGE-Shipin in offline evaluation, since SAGE-Shipin is simultaneously conducting online experiments in Shipin Homepage which makes training dataset already containing SAGE-recommended video transition pairs. Thus a post evaluation of any offline re-trained SAGE model naturally includes echoes to the not-supposed-to-exist user interaction data, hence yielding higher offline metrics than supposed to be. In addition, Deepwalk \citep{Perozzi2014DeepWalkOL} has already been deployed in Shipin Homepage for months, so it is not included in comparisons for the same reason.

\subsection{Offline Evaluation and Results}

\begin{table}
  \caption{Offline model evaluation on multi-scenario datasets. $\mathrm{p}@100$ and $\mathrm{r}@100$ denote $\mathrm{precision}@100$ and $\mathrm{recall}@100$ metrics. Numbers with * are single-scenario results and are not compared directly to MGFN due to mentioned effect.}
  \label{tab:offlineevaluation}
  \begin{tabular}{llcccc}
    \toprule
     & & \multicolumn{2}{c}{Commercial $1$} & \multicolumn{2}{c}{Commercial $2$} \\ \cline{3-6}
     & & $p@100$ & $r@100$ & $p@100$ & $r@100$ \\
    \midrule
    \multirow{3}{*}{\rotatebox[origin=c]{90}{\parbox[c]{1cm}{\centering Single-Scenario}}} & SAGE-Browser & $0.001402$ & $0.01349$ & $0.001280$ & $0.01603$ \\
    & SAGE-Shipin & $0.01561^{\ast}$ & $0.1442^{\ast}$ & $0.01600^{\ast}$ & $0.1559^{\ast}$ \\
    & GAT-Shipin & $0.01575^{\ast}$ & $0.1492^{\ast}$ & $0.01522^{\ast}$ & $0.1478^{\ast}$ \\
    \midrule
    \multirow{5}{*}{\rotatebox[origin=c]{90}{\parbox[c]{1cm}{\centering Multi-Scenario}}} & DataConcat & $0.01398$ & $0.1330$ & $0.01359$ & $0.1304$ \\
    & MGFN-mean & $0.01319$ & $0.1227$ & $0.01423$ & $0.1357$ \\
    & MGFN-weighted & $0.01311$ & $0.1242$ & $0.01372$ & $0.1323$ \\
    & MGFN-GAT & $0.01429$ & $0.1318$ & $0.01362$ & $0.1327$ \\
    & MGFN & $\bm{0.01452}$ & $\bm{0.1339}$ & $\bm{0.01427}$ & $\bm{0.1382}$ \\
    \bottomrule
  \end{tabular}
\end{table}

The offline evaluation result of two commercial datasets is listed in Table~\ref{tab:offlineevaluation}. First of all, MGFN outperforms all other multi-scenario comparisons in cross scenario setting. It confirms the superiority of MGFN in integrating multi-scenario interaction data. Besides, we made following complementary observations:
\begin{itemize}
\item SAGE-Browser vs. SAGE-Shipin: confirms that a disparity in user behavior pattern across scenarios greatly hinders the effectiveness of working models from other scenarios, which is consistent with the exposure bias phenomenon.
\item SAGE-Shipin vs. MGFN: indicates a better performance of SAGE-Shipin over multi-scenario models, however this is expected due to repeatedly strengthened learning on already-exposed items of single scenario models. Hence despite a slight metrics decrease when compared to SAGE-Shipin, MGFN remains competitive over other multi-scenario models and offers extra benefits in improving item variety.
\end{itemize}
\textbf{Component Ablation} In order to achieve an optimal assembly of MGFN and verify individual effects from its components, we ablated various key components of MGFN by comparing to its variants:
\begin{itemize}
\item MGFN vs. MGFN-GAT, SAGE-Shipin vs. GAT-Shipin: ablate graph convolution modules. SAGE-mean aggregator yields a better performance than GAT in general, which we suspect GAT being slightly overfitting on existent interaction data and short of associating non-existent interactions.
\item MGFN vs. MGFN-mean vs. MGFN-weighted: ablates cross-scenario message fusion modules, where concatenation module achieves best offline metrics. 
\end{itemize}

\begin{table}
  \caption{Metrics of negative sampling strategies}
  \label{tab:negsampmetrics}
  \begin{tabular}{l|cc}
    \toprule
    Strategy & $\mathrm{p}@100$ & $\mathrm{r}@100$ \\
    \midrule
    Degree negative in Shipin & $0.000131$ & $0.0001664$ \\
    Degree negative in Browser & $0.002478$ & $0.0171$ \\
    Random negative in both scenarios & $0.01054$ & $0.1023$ \\
    Cross-scenario negative sampling & $0.01452$ & $0.1339$ \\
    \bottomrule
  \end{tabular}
\end{table}

\textbf{Negative Sampling Strategy} Since a choice on negative sampling strategy is essential to lifting the upper limit of algorithms, we investigate MGFN performance under $4$ mentioned negative sampling strategies. The result is listed in Table ~\ref{tab:negsampmetrics}, showing that MGFN equipped with cross-scenario negative sampling outperforms the others, and degree negative sampling in one single scenario actually biases MGFN with a narrowed set of negative samples. 

\subsection{Online A/B Test}

We investigate model online performance in production A/B environments. Three models including base, single-scenario SAGE-Shipin and a multi-scenario model (DataConcat or MGFN) share one experiment bucket and are assigned a customer flow of $30\%, 35\%, 35\%$ respectively. Note that online metrics CTR, VV and VWT do not rely on the amount of exposure. 

\begin{table}
  \caption{Online results of first $10$-day average after deployment. Phase number denotes different periods of time.}
  \label{tab:onlineab}
  \begin{tabular}{lcccc}
    \toprule
    Phase $\uppercase\expandafter{\romannumeral1}$ & User Type & CTR & VV & VWT \\
    \midrule
    \multirow{3}{1.5cm}{DataConcat} & all users & $+0.35\%$ & $+0.29\%$ & $-0.27\%$ \\
     & active users & $+0.32\%$ & $+0.34\%$ & $-0.26\%$ \\
     & less-active users & $+0.57\%$ & $+0.29\%$ & $+0.13\%$ \\
    \midrule
    \multirow{3}{1.5cm}{SAGE-Shipin} & all users & $+0.40\%$ & $+0.07\%$ & $-0.45\%$ \\
     & active users & $+0.47\%$ & $+0.18\%$ & $-0.33\%$ \\
     & less-active users & $+0.16\%$ & $+0.12\%$ & $-0.56\%$ \\
    \bottomrule
  \end{tabular}  
  
  \begin{tabular}{lcccc}
    \toprule
    Phase $\uppercase\expandafter{\romannumeral2}$ & User Type & CTR & VV & VWT \\
    \midrule
    \multirow{3}{1.5cm}{MGFN} & all users & $+0.06\%$ & $+0.47\%$ & $-0.01\%$ \\
     & active users & $+0.00\%$ & $+0.54\%$ & $+0.04\%$ \\
     & less-active users & $\bm{+0.63\%}$ & $\bm{+0.71\%}$ & $\bm{+0.31\%}$ \\
    \midrule
    \multirow{3}{1.5cm}{SAGE-Shipin} & all users & $+0.21\%$ & $+0.80\%$ & $+0.02\%$ \\
     & active users & $+0.14\%$ & $+0.87\%$ & $+0.05\%$ \\
     & less-active users & $\bm{+0.87\%}$ & $\bm{+1.00\%}$ & $\bm{+0.49\%}$ \\
    \bottomrule
  \end{tabular}  
  
\end{table}

The first $10$-day average online result of SAGE-Shipin and MGFN (or DataConcat) are listed in Table~\ref{tab:onlineab}. The precise metric values are not exposed for commercial concerns, and numbers are displayed as ratio to "base". Users are separated into two groups based on their use frequency of Shipin, where users who have watched less than $20$ videos in the past $7$ days (including new users who have absolutely not used Shipin before) are denoted "new users", and "old users" otherwise. It's observed that MGFN demonstrated favorable performance on new users, though not exceeding single-scenario SAGE, still validates MGFN's overall advantage over deployed base.

\textbf{Outer-scenario Preference Metrics} Apart from overall metrics, the value of MGFN lies on its extraordinary ability in increasing item variety. Here, we utilize \textit{Outer-scenario Preference Metrics} to quantify video variety change. It includes two metrics: \textit{number of outer-scenario video watches} (i.e. total video watches of Browser-exclusive videos which in the past $7$-day have not been watched in Shipin), and \textit{number of unique outer-scenario videos} imported to the current scenario (i.e. number of watched Browser-exclusive videos). In a single-scenario setting, the numbers are $0$. When "base" remains unchanged, the change of \textit{number of outer-scenario video watches} is able to reveal user preference on Browser-exclusive videos against Shipin videos, and an increase in \textit{number of unique outer-scenario videos} also indicates the amount of activated cold videos in Shipin scenario (since videos are supplied by a same reservoir).

\begin{figure}[htb]
  \centering
  \subfigure[Number of outer-scenario video watches]{
	\includegraphics[width=0.8\columnwidth]{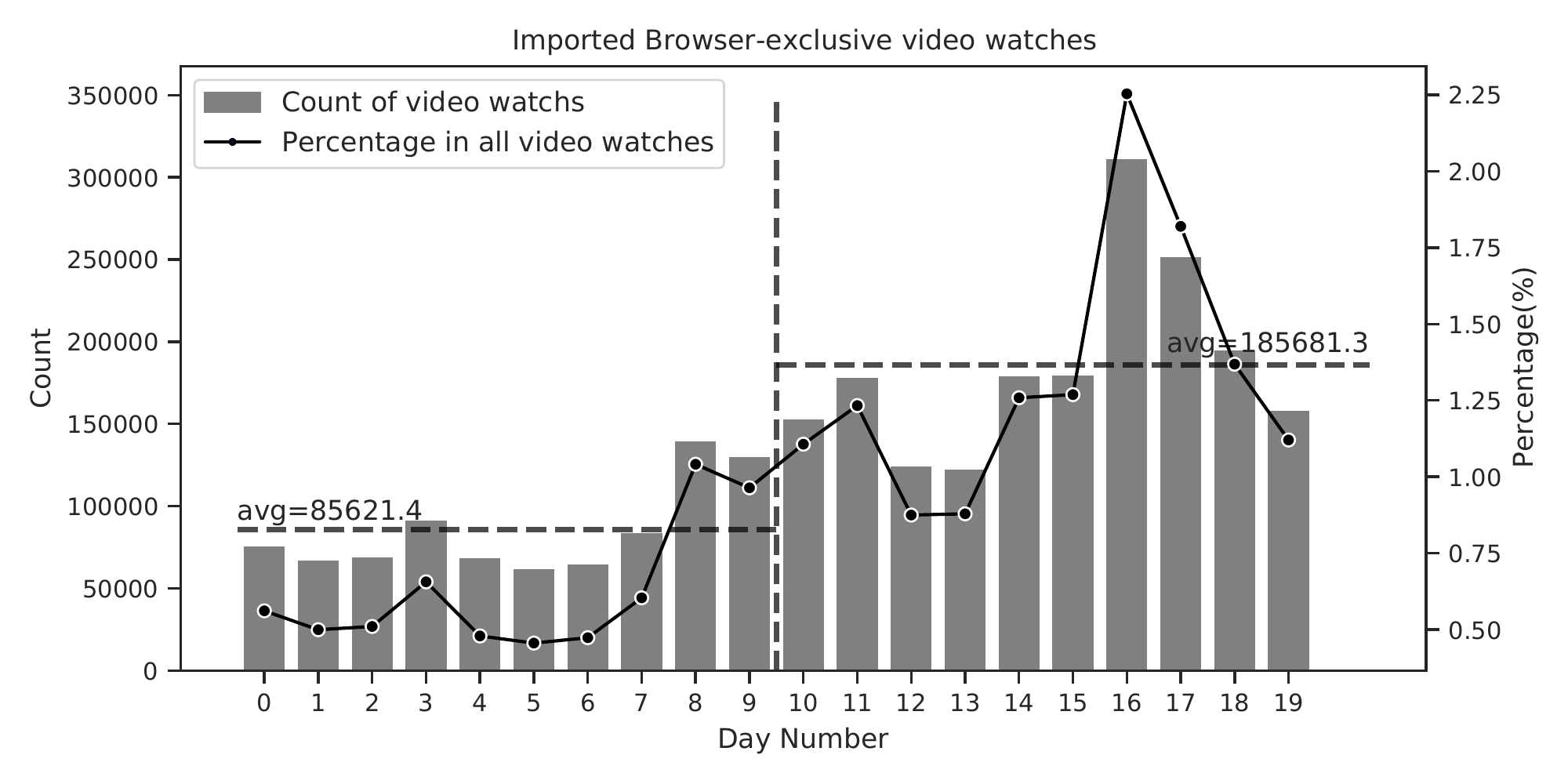}
  }
  \subfigure[Number of outer-scenario videos]{
	\includegraphics[width=0.8\columnwidth]{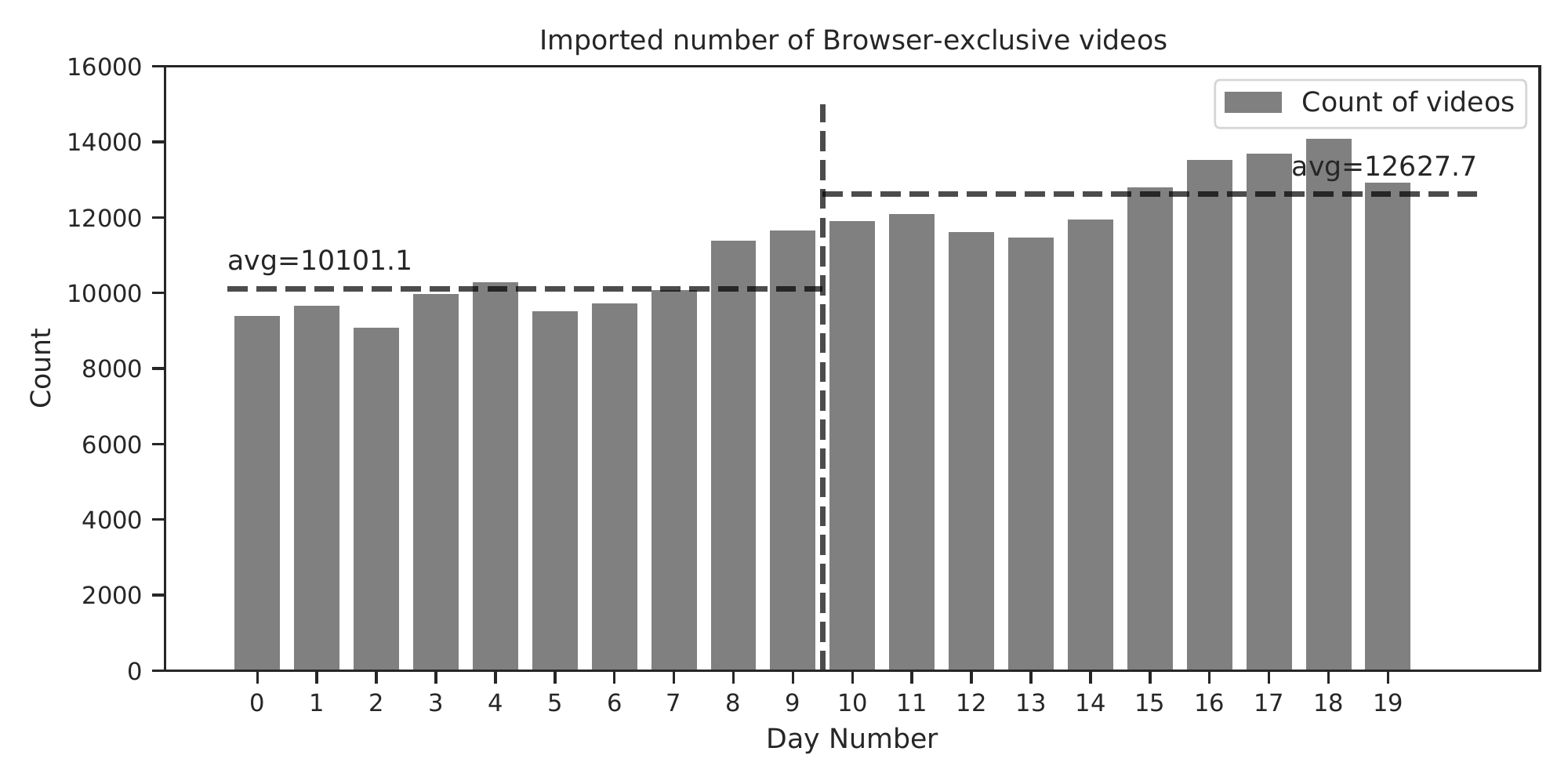}
  }
  \caption{Change of Outer-Scenario Preference Metrics since deployment. These two metrics demonstrate a great success of MGFN against DataConcat in increasing cross-scenario video watches by $116\%$ and activating cold videos by $25\%$. For single-scenario models, the numbers are $0$.}
  \Description{The top figure is entitled "Imported Browser-exclusive video watches". The horizontal axis is labeled "Day Number" and ranges from $0$ to $19$. There are a bar plot labeled "Count of video watches", and a line plot labeled "Percentage in all video watches". The vertical axis for bar plot is labeled "Count" and ranges from $0$ to $350000$, while the vertical axis for line plot is labeled "Percentage ($\%$)" and ranges from $0.50$ to $2.25$. The graph is split into two periods of time, days $0$ to $9$ and days $10$ to $19$. The left $10$ days are labeled with "avg=$85621.4$" video watches, and the right $10$ days are labeled with "avg=$185681.3$" video watches. The bottom figure is a bar plot entitled "Imported number of Browser-exclusive videos". The horizontal axis is labeled "Day Number" and ranges from $0$ to $19$, and the vertical axis is labeled "Count" ranging from $0$ to $16000$. The graph is split into two periods of time, days $0$ to $9$ and days $10$ to $19$. The left $10$ days are labeled with "avg=$10101.1$" videos, and the right $10$ days are labeled with "avg=$12627.7$" videos.}
  \label{outerscenariometrics}
\end{figure}

We sum up \textit{outer-scenario preference metrics} since deployment, and the results are shown in Fig.~\ref{outerscenariometrics}. When MGFN compared to DataConcat in a $10$-day period, \textit{number of outer-scenario video watches} increases from $86621.4$ to $185681.3$, and \textit{number of unique outer-scenario videos} increases from $10101.1$ to $12627.7$, demonstrating an average increase of $\bm{116\%}$ and $\bm{25\%}$ respectively. The "imported" Browser-exclusive videos have not been watched in Shipin Homepage, but MGFN is able to collect their recent watch record in Browser Feeds and feed them again into Shipin. This action simply reuses those less-watched videos and forms a virtuous cycle across scenarios, which alleviates both exposure bias and cold-start problem at once. As a result, we assured MGFN's strong capability in increasing recommendation variety.

\subsection{Discussion}

\subsubsection{Visualization of learnt video embeddings}

\begin{figure}[htb]
  \centering
  \includegraphics[width=\columnwidth]{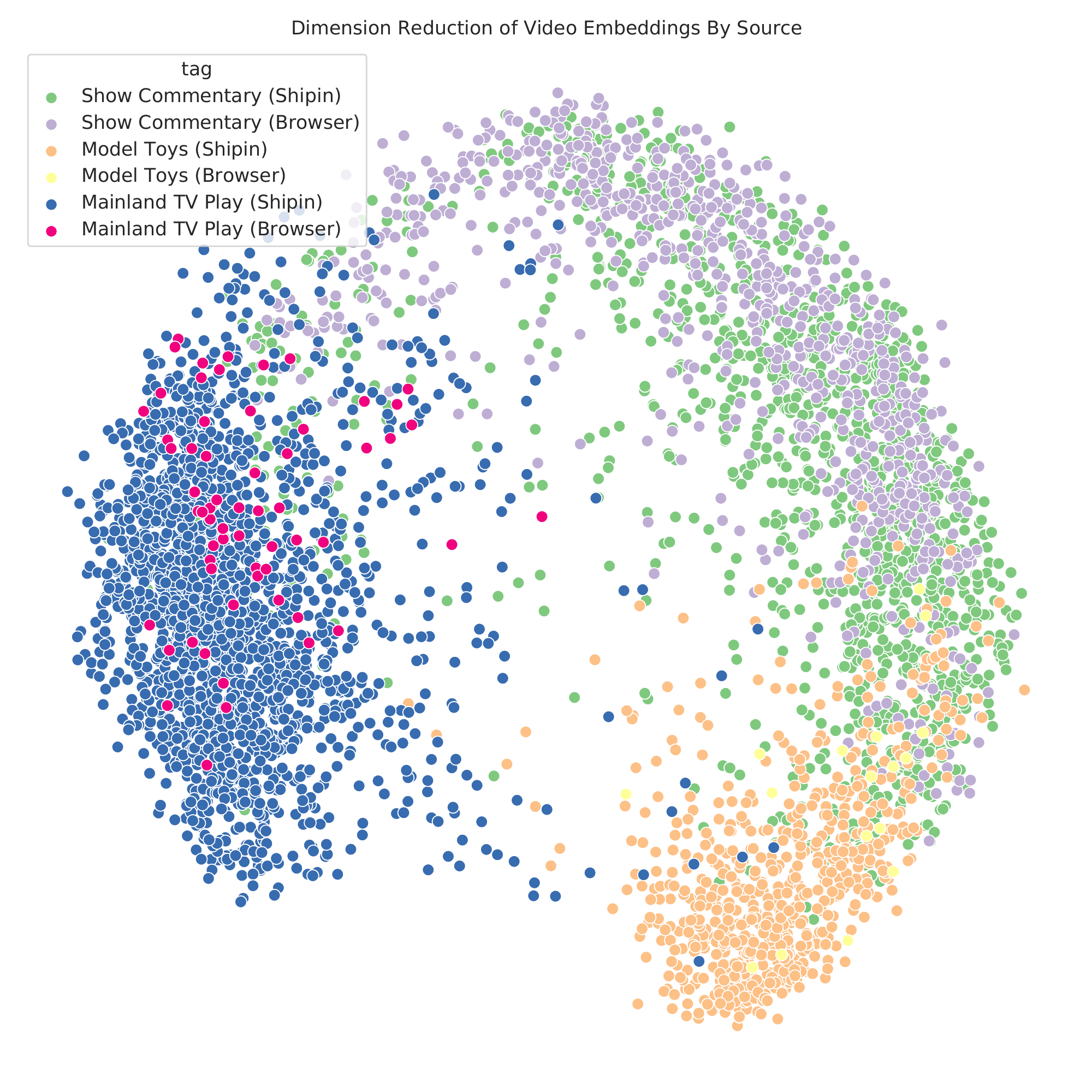}
  \caption{PCA decomposition of selected cross-scenario video embeddings (better viewed in color).}
  \Description{The graph is a scattered plot entitled "Dimension Reduction of Video Embeddings by Source". There are six types of colored dots indicating different tags, blue dots refers to "Show Commentary (Shipin)", purple dots refers to "Show Commentary (Browser)", orange dots refers to "Model Toys (Shipin)", yellow dots refers to "Model Toys (Browser)", blue dots refers to "Mainland TV Play (Shipin)", and red dots refers to "Mainland TV Play (Browser)". Dots with tags from same category are located nearby.}
  \label{fig:pca}
\end{figure}

The question remains that DO learnt video embeddings truely demonstrate physical meanings? We randomly select $5000$ videos from top $3$ tags and perform PCA dimension reduction on their learnt embeddings. The plot (as in Fig.~\ref{fig:pca}) shows that, even coming from different scenarios, videos with similar tags are still projected into closer positions in the embedding space. The major reason is that MGFN projects topologically-linked items into affinity positions, thus chaining videos across scenarios. Upon deployment, chained videos can be easily retrieved via nearest neighbor search.

\subsubsection{Further applications} 

In our experiments, MGFN demonstrates a great success in marking up less-watched videos via linking scenario-variant preferences. Conceptual further applications would include recommending popular products overseas across e-commerce submarkets, or mining inactive posts in niche forums.

\subsubsection{Privacy concerns \label{privacyconcerns}}

Although conditional inner-app data sharing is legal, distributing user-sensitive data out of individual apps is strictly forbidden by law. For our platform mounting thousands of individual applications, regular legal approaches (e.g. sharing anonymized non-sensitive behavior-related data) would disable user-level cold-start, and it would be an overall uncomfortable experience for users to discover past search/watch histories from similar applications, therefore this work only focuses on item-level global representation learning.

\section{Conclusion}

This paper addresses a multi-graph structured multi-scenario recommendation solution to alleviate both exposure bias and cold-start problem. We constructed Cross-Scenario Multi-Graph to encapsulate interaction patterns across scenarios, and explored its working mechanism in detail. We proposed a novel Multi-Graph Fusion Networks along with a specific cross-scenario negative sampling strategy to capture topological patterns. We validated model performance through offline and online experiments in production setting. We verified MGFN's strong ability in alleviating exposure bias and activating cold videos. Future improvements could be expected in the field of time-evolving graph representation learning and real-time training in online services.

\begin{acks}
We would like to thank OPPO Content Recommendation Group and OPPO Finance Center AI Group for their warm and continuous support, and our group members for discussion and generous help. We would also like thank all anonymous reviewers for precious suggestions and comments.
\end{acks}

\bibliographystyle{ACM-Reference-Format}
\bibliography{mgfn}

\appendix

\section{Training Details}

The MGFN model is trained with the help of PyTorch-based DGL framework \citep{wang2019dgl}. The minibatches of sampled triplets and their $2$-hop neighborhood with determined number of neighbors are sampled in CPU with $24$ multi-processing threads and are transmitted to one GPU for model training. With a time span of $7$ days, the offline training process (including data preprocessing, graph construction and model training) takes roughly $4.5$ hours in total. 

\section{Model Parameters}

For all aforementioned models, the learnt video embedding size is set to $128$. Input node embedding after feature transformation is $128$d. Each positive transition is conjugated with $5$ negatives in negative sampling, and batch size is chosen as $4000$. In order to fully train the models, dropout is chosen from $\{0.0, 0.5\}$, learning rate is selected from $\{0.001, 0.002, 0.005\}$, and models are trained with stop steps ranging from $10000$ to $16000$. The baseline models are initialized with mentioned parameters but only the optimal result is reported in the table.

\end{document}